\documentclass{article}

\usepackage[margin=1.0in]{geometry}

\usepackage[utf8]{inputenc} 
\usepackage[T1]{fontenc}    
\usepackage{hyperref}       
\usepackage{url}            
\usepackage{booktabs}       
\usepackage{amsfonts}       
\usepackage{nicefrac}       
\usepackage{microtype}      
\usepackage{acronym}
\usepackage{amsmath, amsthm, amssymb, amsfonts, mathtools}
\usepackage[english]{babel}
\usepackage{sidecap}
\usepackage{xcolor}
\usepackage{graphicx}
\graphicspath{{img/}}
\usepackage[T1]{fontenc}
\usepackage{tablefootnote}
\usepackage{units}
\usepackage{verbatim}
\usepackage{times}
\usepackage{mathtools}
\usepackage{algorithm, algorithmic}

\renewcommand{\refeq}[1]{{Eq.~(\ref{#1})}}

\newcommand{\reffig}[1]{{\color{blue!70}(Fig.~\ref{#1})}}
\newcommand{\reftab}[1]{{\color{blue!70}(Tab.~\ref{#1})}}

\newcommand{\refsec}[1]{{Sec.~\ref{#1}}}

\normalsize

\title{Synaptic Plasticity Dynamics for Deep Continuous Local Learning (DECOLLE)}

\author{%
Jacques Kaiser,\\
 FZI Research Center For Information Technology\\ Karlsruhe, Germany \\
 ~\\
 Hesham Mostafa, \\
 Department of Bioengineering\\ University of California San Diego\\ La Jolla, USA \\
 ~\\
 Emre Neftci, \\
 Department of Cognitive Sciences\\ Department of Computer Science\\ University of California Irvine, Irvine, USA \\
}

\begin{document}

\acrodef{AC}[AC]{Arrenhius \& Current}
\acrodef{AD}[AD]{Automatic Differentiation}
\acrodef{ANN}[ANN]{Artificial Neural Network}
\acrodef{AER}[AER]{Address Event Representation}
\acrodef{AEX}[AEX]{AER EXtension board}
\acrodef{AMDA}[AMDA]{``AER Motherboard with D/A converters''}
\acrodef{API}[API]{Application Programming Interface}
\acrodef{BP}[BP]{Back-Propagation}
\acrodef{BPTT}[BPTT]{Back-Propagation-Through-Time}
\acrodef{BM}[BM]{Boltzmann Machine}
\acrodef{CAVIAR}[CAVIAR]{Convolution AER Vision Architecture for Real-Time}
\acrodef{CCN}[CCN]{Cooperative and Competitive Network}
\acrodef{CD}[CD]{Contrastive Divergence}
\acrodef{CMOS}[CMOS]{Complementary Metal--Oxide--Semiconductor}
\acrodef{COTS}[COTS]{Commercial Off-The-Shelf}
\acrodef{CPU}[CPU]{Central Processing Unit}
\acrodef{CV}[CV]{Coefficient of Variation}
\acrodef{CV}[CV]{Coefficient of Variation}
\acrodef{DAC}[DAC]{Digital--to--Analog}
\acrodef{DBN}[DBN]{Deep Belief Network}
\acrodef{DCLL}[DECOLLE]{Deep Continuous Local Learning}
\acrodef{DFA}[DFA]{Deterministic Finite Automaton}
\acrodef{DFA}[DFA]{Deterministic Finite Automaton}
\acrodef{divmod3}[DIVMOD3]{divisibility of a number by 3}
\acrodef{DPE}[DPE]{Dynamic Parameter Estimation}
\acrodef{DPI}[DPI]{Differential-Pair Integrator}
\acrodef{DSP}[DSP]{Digital Signal Processor}
\acrodef{DVS}[DVS]{Dynamic Vision Sensor}
\acrodef{EDVAC}[EDVAC]{Electronic Discrete Variable Automatic Computer}
\acrodef{EIF}[EI\&F]{Exponential Integrate \& Fire}
\acrodef{EIN}[EIN]{Excitatory--Inhibitory Network}
\acrodef{EPSC}[EPSC]{Excitatory Post-Synaptic Current}
\acrodef{EPSP}[EPSP]{Excitatory Post--Synaptic Potential}
\acrodef{eRBP}[eRBP]{Event-Driven Random Back-Propagation}
\acrodef{FPGA}[FPGA]{Field Programmable Gate Array}
\acrodef{FSM}[FSM]{Finite State Machine}
\acrodef{GPU}[GPU]{Graphical Processing Unit}
\acrodef{HAL}[HAL]{Hardware Abstraction Layer}
\acrodef{HH}[H\&H]{Hodgkin \& Huxley}
\acrodef{HMM}[HMM]{Hidden Markov Model}
\acrodef{HW}[HW]{Hardware}
\acrodef{hWTA}[hWTA]{Hard Winner--Take--All}
\acrodef{IF2DWTA}[IF2DWTA]{Integrate \& Fire 2--Dimensional WTA}
\acrodef{IF}[I\&F]{Integrate \& Fire}
\acrodef{IFSLWTA}[IFSLWTA]{Integrate \& Fire Stop Learning WTA}
\acrodef{INCF}[INCF]{International Neuroinformatics Coordinating Facility}
\acrodef{INI}[INI]{Institute of Neuroinformatics}
\acrodef{IO}[IO]{Input-Output}
\acrodef{IPSC}[IPSC]{Inhibitory Post-Synaptic Current}
\acrodef{ISI}[ISI]{Inter--Spike Interval}
\acrodef{JFLAP}[JFLAP]{Java - Formal Languages and Automata Package}
\acrodef{LIF}[LI\&F]{Leaky Integrate \& Fire}
\acrodef{LSM}[LSM]{Liquid State Machine}
\acrodef{LSTM}[LSTM]{Long Short Term Memory}
\acrodef{LTD}[LTD]{Long-Term Depression}
\acrodef{LTI}[LTI]{Linear Time-Invariant}
\acrodef{LTP}[LTP]{Long-Term Potentiation}
\acrodef{LTU}[LTU]{Linear Threshold Unit}
\acrodef{MCMC}{Markov Chain Monte Carlo}
\acrodef{NHML}[NHML]{Neuromorphic Hardware Mark-up Language}
\acrodef{NMDA}[NMDA]{NMDA}
\acrodef{NME}[NE]{Neuromorphic Engineering}
\acrodef{PCB}[PCB]{Printed Circuit Board}
\acrodef{PRC}[PRC]{Phase Response Curve}
\acrodef{PSC}[PSC]{Post-Synaptic Current}
\acrodef{PSP}[PSP]{Post--Synaptic Potential}
\acrodef{RI}[KL]{Kullback-Leibler}
\acrodef{RRAM}[RRAM]{Resistive Random-Access Memory}
\acrodef{RTRL}[RTRL]{Real-Time Recurrent Learning}
\acrodef{RBM}[RBM]{Restricted Boltzmann Machine}
\acrodef{ROC}[ROC]{Receiver Operator Characteristic}
\acrodef{SAC}[SAC]{Selective Attention Chip}
\acrodef{SCD}[SCD]{Spike-Based Contrastive Divergence}
\acrodef{SCX}[SCX]{Silicon CorteX}
\acrodef{SRM}[SRM$_0$]{Spike Response Model}
\acrodef{SNN}[SNN]{Spiking Neural Network}
\acrodef{STDP}[STDP]{Spike Time Dependent Plasticity}
\acrodef{SW}[SW]{Software}
\acrodef{sWTA}[SWTA]{Soft Winner--Take--All}
\acrodef{VHDL}[VHDL]{VHSIC Hardware Description Language}
\acrodef{VLSI}[VLSI]{Very  Large  Scale  Integration}
\acrodef{WTA}[WTA]{Winner--Take--All}
\acrodef{XML}[XML]{eXtensible Mark-up Language}

\maketitle

\begin{abstract}
A growing body of work underlines striking similarities between biological neural networks and recurrent, binary neural networks. A relatively smaller body of work, however, addresses the similarities between learning dynamics employed in deep artificial neural networks and synaptic plasticity in spiking neural networks.
The challenge preventing this is largely caused by the discrepancy between the dynamical properties of synaptic plasticity and the requirements for gradient backpropagation.
Learning algorithms that approximate gradient backpropagation using local error functions can overcome this challenge.
Here, we introduce \ac{DCLL}, a spiking neural network equipped with local error functions for online learning with no memory overhead for computing gradients.
DECOLLE is capable of learning deep spatiotemporal representations from spikes relying solely on local information, making it compatible with neurobiology and neuromorphic hardware.
Synaptic plasticity rules are derived systematically from user-defined cost functions and neural dynamics by leveraging existing autodifferentiation methods of machine learning frameworks.
We benchmark our approach on the event-based neuromorphic dataset N-MNIST and DvsGesture, on which DECOLLE performs comparably to the state-of-the-art.
DECOLLE networks provide continuously learning machines that are relevant to biology and supportive of event-based, low-power computer vision architectures matching the accuracies of conventional computers on tasks where temporal precision and speed are essential.
\end{abstract}
\section{Introduction}

Understanding how the plasticity dynamics in multilayer biological neural networks are organized for efficient data-driven learning is a long-standing question in computational neurosciences \cite{Zenke_Ganguli17_supesupe,Sussillo_Abbott09_genecohe,Clopath_etal10_connrefl}.
The generally unmatched success of deep learning algorithms in a wide variety of data-driven tasks prompts the question of whether the ingredients of their success are compatible with their biological counterparts, namely \acp{SNN}.
Biological neural networks distinguish themselves from \acp{ANN} by their continuous-time dynamics, the locality of their operations \cite{Baldi_etal17_learmach}, and their spike(event)-based communication.
Taking these properties into account in a neural network is challenging, as the spiking nature of the neurons' nonlinearity makes it non-differentiable, the continuous-time dynamics raise a temporal credit assignment problem and the assumption of computations being local to the neuron disqualifies the use of \ac{BPTT}.

\begin{figure}
\begin{center}
  \includegraphics[height=.33\textheight]{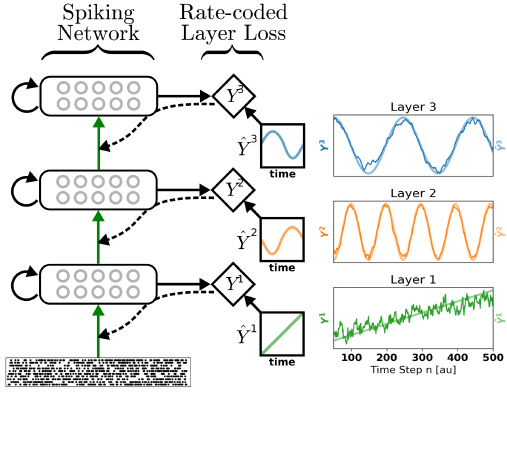}
  \hspace{.02\textwidth}
  \includegraphics[height=.33\textheight]{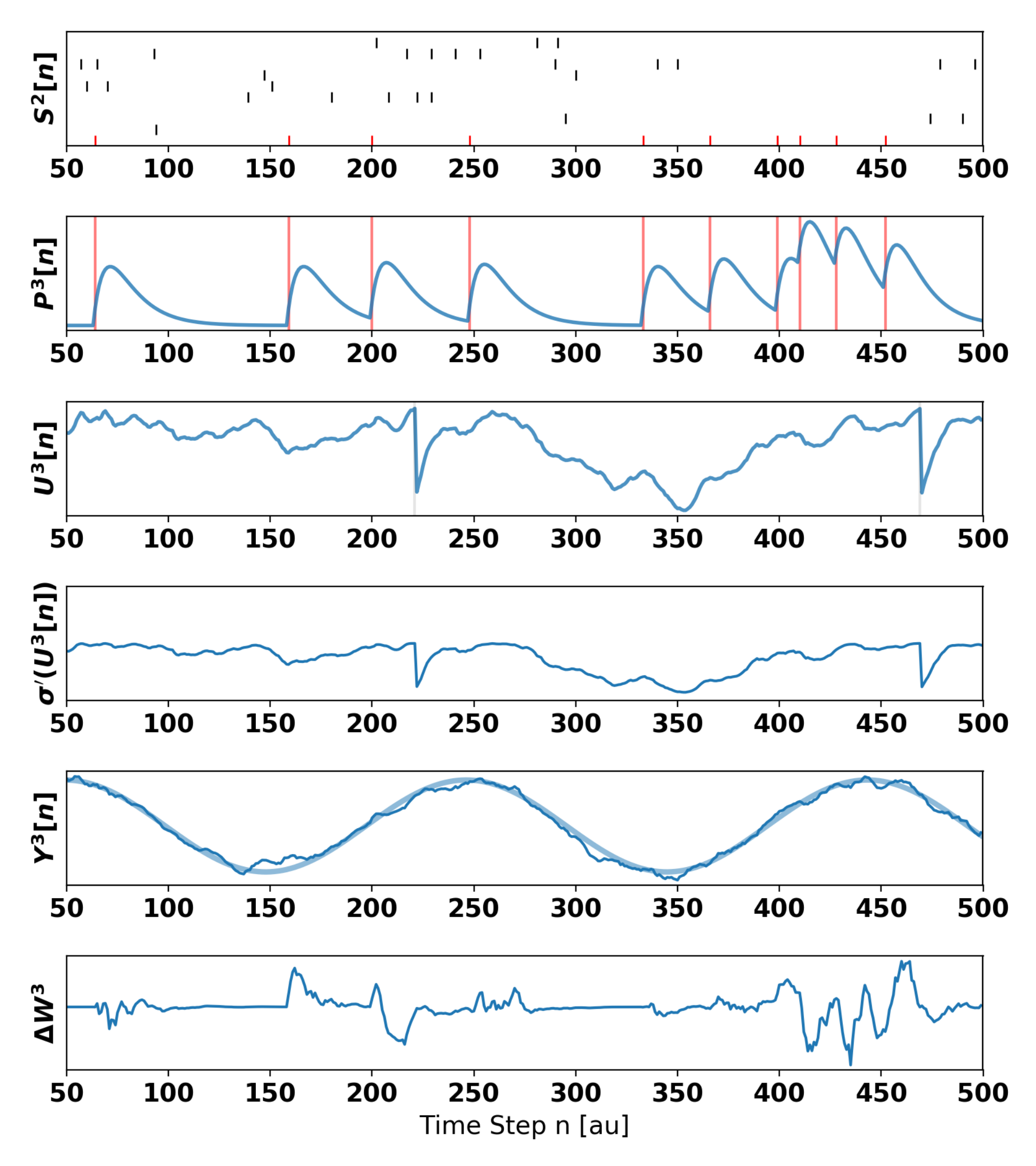}
\end{center}
\caption{\label{fig:dcll_illustration}\acf{DCLL}. (Left) Each layer consists of spiking neurons with continuous dynamics. Each layer feeds into a local readout units through fixed, random connections (diamond-shaped, $y$). The local layer is trained such that the readout at each layer produce auxiliary targets $\hat{Y}$. Errors are propagated through the random connections to train weights coming into the spiking layer, but no further (curvy, dashed line). To simplify the learning rule and enable linear scaling of the computations, the cost function is a function of the states in the same time step. The state of the spiking neurons (membrane potential, synaptic states, refractory state) is carried forward in time. Consequently, even in the absence of recurrent connections, the neurons are stateful in the sense of recurrent neural networks.
(Right) Snapshot of the neural states illustrating the \ac{DCLL} learning rule in the top layer. In this example, the network is trained to produce three time-varying pseudo-targets $\hat{Y}^1$, $\hat{Y}^2$, $\hat{Y}^3$.}
\end{figure}

In this article, we describe \ac{DCLL}, a \ac{SNN} model with plasticity dynamics that solves the three problems above, and that performs at proficiencies comparable to that of multilayer neural networks.
\ac{DCLL} uses layerwise local readouts \cite{Mostafa_etal17_deepsupe}, which enables gradients to be computed locally \reffig{fig:dcll_illustration}.
To tackle the temporal dynamics of the neurons, we use a recently established equivalence between \acp{SNN} and recurrent \acp{ANN} \cite{Neftci_etal19_surrgrad}.
This equivalence rests on a computational graph of the \ac{SNN}, which can be implemented with standard machine learning frameworks as a recurrent neural network.
Unlike \ac{BPTT} and like \ac{RTRL}\cite{Williams_Zipser89_learalgo}, \ac{DCLL} is formulated in a way that the information necessary to compute the gradient is propagated forward, making the plasticity rule temporally local.
Existing rules of this sort require dedicated state variables for every synapse, thus scaling at least quadratically with the number of neurons \cite{Williams_Zipser89_learalgo,Zenke_Ganguli17_supesupe}.
In contrast, \ac{DCLL} scales linearly with the number of neurons.
This is achieved using a spatially and temporally local cost function reminiscent of readout mechanisms used in liquid state machines \cite{Maass_etal02_realcomp}, but where the readout is performed over a fixed random combination of the neuron outputs.
Our approach can be viewed as a type of synthetic gradient, a technique used to decouple one or more layers from the rest of the network to prevent layerwise locking \cite{Jaderberg_etal16_deconeur}.
Although synthetic gradients usually involve an outer loop that is equivalent to a full \ac{BP} through the network, \ac{DCLL} instead relies on the random initialization of the local readout and forgoes the outer loop.

Conveniently, \ac{DCLL} can leverage existing autodifferentiation tools of modern machine learning frameworks.
Its linear scalability enables the training of hundreds of thousands of spiking neurons on a single GPU, and continual learning on very fine time scales.
We demonstrate our approach on the classification of gestures, the IBM DvsGesture dataset \cite{Amir_etal17_lowpowe}, recorded using an event-based neuromorphic sensor and report comparable performance to deep neural networks and even networks trained with \ac{BPTT}.

\subsection{Related Work}
Previous work demonstrated learning in multiple layers of \ac{SNN} using feedback alignment \cite{Neftci_etal17_evenranda,Lillicrap_etal16_randsyna}, performing at about $2\%$ classification error on MNIST.
However, those networks operated in the firing rate regime, by using either large populations or slow dynamics.
In those works, training was not insensitive to the temporal dynamics of the neurons.
The need for temporal dynamics are often obfuscated by the static nature of the benchmarked problems (\emph{e.g.} MNIST), and a long readout interval that allows to ignore initial transients caused by the dynamics.
In our previous work \cite{Neftci_etal17_evenranda}, ignoring temporal dynamics raised a ``loop duration'' problem, \emph{i.e.} that the errors are available only after they have propagated through the network. 
This introduces latency or requires additional buffers for storing intermediate neural states.
In traditional deep learning, the loop duration manifests itself as ``layerwise locking'', during which a layer's weights cannot be updated until a global cost function is evaluated \cite{Jaderberg_etal16_deconeur}.
This causes underutilization of the computing resources and a slowdown in learning.
Besides the loop duration problem, multilayer networks trained with feedback alignment cannot reach the performances of gradient \ac{BP}, especially with deeper networks ($\ge 30\%$ accuracy drop on ImageNet compared to backpropagation \cite{Bartunov_etal18_assescal}).

The complex dynamics of spiking neurons is an important feature that can be exploited for learning spatiotemporal patterns.
In a single layer of neurons, this feature can be leveraged using gradient descent, since it is applicable to the subthreshold dynamics of leaky \ac{IF} neurons \cite{Gutig_Sompolinsky06_tempneur,Bohte_etal00_spikback}. 
Because the \ac{IF} neuron output is non-differentiable, however, the application of these approaches to muliple layers is not straightforward.
To deal with this problem, SuperSpike uses a surrogate network with differentiable activation functions to compute an approximate gradient \cite{Zenke_Ganguli17_supesupe}.
The authors show that this learning rule is equivalent to a forward-propagation of errors using synaptic traces, and is capable of learning in hidden layers of feedforward multilayer networks.
%

Because the traces need to be computed for every trainable parameter, Superspike scales temporally and spatially as $O(N^2)$, where $N$ is the number of neurons.
While the complex biochemical processes at the synapse could account for the quadratic scaling, it prevents an efficient implementation in available hardware.
Like SuperSpike, \ac{DCLL} uses surrogate gradients to perform weight updates, but as discussed later, the cost function is local in time and space, such that only one trace per input neuron is required.
This enables the algorithm to scale linearly in space.
Furthermore, in \ac{DCLL} the computation of the gradients can reuse the variables computed for the forward dynamics, such that learning has no additional memory overhead.

\ac{DCLL} has some resemblance with reservoir networks, which are neural networks with fixed internal connectivity and trainable readout functions  \cite{Jaeger01_echostat,Eliasmith_Anderson04_neurengi,Maass_etal02_realcomp,Sussillo_Abbott09_genecohe}. 
The local readout in \ac{DCLL} acts like a decoder layer in the flavor of the linear readouts in reservoir networks.
In contrary to reservoir networks, \ac{DCLL} learns the internal weights, but the readout weights are random and fixed. 
The training of the internal weights allows the network to learn representations that are easier to classify inputs for subsequent layers \cite{Mostafa_etal17_deepsupe}.

Spiking neural networks can be viewed as a subclass of binary, recurrent \acp{ANN} \cite{Neftci_etal19_surrgrad}.
In the \ac{ANN} sense, they are recurrent even when all the connections are feed-forward because the neurons maintain a state that is propagated forward at every time step.
Binary neural networks, where both activations and/or weights are binary were studied in deep learning as a way to decrease model complexity during inference \cite{Courbariaux_etal16_binaneur,Rastegari_etal16_xnorimag}.
\ac{BPTT} for training \acp{SNN} was investigated in \cite{Bohte_etal00_spikback,Lee_etal16_traideep,Shrestha_Orchard18_slayspik,Bellec_etal18_longshor,Huh_Sejnowski17_graddesc}.
\ac{BPTT}-based approaches provide an unbiased estimation of the gradients but at a cost in memory, because the entire sequence and resulting activity states are stored to compute gradients.
Although the truncation of the sequences (as in truncated \ac{BPTT}) can mitigate this problem, it is not adequate when discretizing continuous-time networks, such as the \ac{SNN} \cite{Neftci_etal19_surrgrad} because the sequences can consists of hundreds of steps.
This is because the time constants and simulation timestep in \acp{SNN} are such that the truncation window must be much larger. 
For \ac{SNN} simulations with biological time constants, it is common to use simulation time steps $\Delta t \le 1 \mathrm{ms}$.
Smaller time steps capture non-linear dynamics more accurately and determine the temporal precision of all produced spike times.
Assuming $\Delta t = 1 \mathrm{ms}$ (as used in this work), and if relevant interactions occur at one second, this implies that the truncation window must be at about 1000 timesteps.
This significantly increases the complexity of \ac{BPTT} in \acp{SNN}.
In practice, the size of \ac{SNN} trainable by \ac{BPTT} is severely limited by the available GPU memory \cite{Shrestha_Orchard18_slayspik}.
As we explain later in this article, \ac{DCLL} requires an order $T$ less memory resources compared to \ac{BPTT}, where $T$ is the sequence length.
Hence, \ac{DCLL} networks are generally not memory-limited.
Furthermore, \ac{DCLL} can be formulated as a local, three-factor synaptic plasticity rule, and is thus amenable to implementation in dedicated, event-based (neuromorphic) hardware \cite{Davies_etal18_loihneur} and compatible with neurobiology.

Decoupled Neural Interfaces (DNI) were proposed to mitigate layerwise locking in training deep neural networks \cite{Jaderberg_etal16_deconeur}.
In DNIs, this decoupling is achieved using a synthetic gradient, a neural network that estimates the gradients for a portion of the network.
In an inner loop, the network parameters are trained using the synthetic gradients, and in an outer loop, the synthetic gradient network parameters are trained using a full \ac{BP} step.
The gradient computed using local errors in \ac{DCLL} described below can be viewed as a type of synthetic gradient, which ignores the outer loop to avoid a full \ac{BP} step.
Although ignoring the outer loop limits \ac{DCLL}'s adaptation of the features using errors from other layers, we find that the network performs at or above state-of-the-art accuracy on N-MNIST and DVS Gesture benchmark tasks.

A related method called E-prop was developed in parallel to \ac{DCLL} \cite{Bellec_etal19_biolinsp}.
The resulting learning rule in E-prop is of the same form as Superspike and \ac{DCLL}.
E-prop uses adaptive spiking \ac{LSTM} neurons to maintain a longer term memory.
This generalization allows to solve tasks with long-term dependencies (similar to \acp{LSTM}) but requires maintaining one trace per synapse.
These memory requirements quickly exceed the capabilities of modern GPUs, especially when applied to convolutional neural networks.
Even in neuromorphic hardware, maintaining a synapse-specific trace can incur a prohibitive cost in area and power \cite{Huayaney_etal16_learsili,Davies_etal18_loihneur}.
In \ac{DCLL}, we focus on networks which do not incur any memory overhead for training, allowing to tractably train large networks.

This work builds on a combination of how gradients are dynamically computed in SuperSpike and local errors. We show in the methods section that this combination considerably reduces the computational requirements compared to a computing a global loss.

\section{Methods}

\subsection{Neuron and Synapse Model}
The neuron and synapse models used in this work follow leaky, current-based \ac{IF} dynamics with a relative refractory mechanism.
The dynamics of the membrane potential $U_i$ of a neuron $i$ is governed by the following differential equations:
\begin{equation}\label{eq:clif}
  \begin{split}
    U_i(t) =& V_i(t) - \rho R_i(t) + b_i,\\
    \tau_{mem}\frac{\mathrm{d}}{\mathrm{d}t} V_i(t) =& - V_i(t) + I_i(t),\\
    \tau_{ref} \frac{\mathrm{d}}{\mathrm{d}t} R_i(t) =& -R_i(t)  +  S_i(t),\\
  \end{split}
\end{equation}
\noindent with $S_i(t)$ the binary value ($0$ or $1$) representing whether neuron $i$ spiked at time $t$.
The separation of the membrane potential into two variables $U$ and $V$ is done here for implementations reasons only.
Biologically, the two states can be interpreted as a special case of a two-compartment model, consisting of one dendritic ($V$) and one somatic ($U$) compartment \cite[Chap. 6.4]{Gerstner_etal14_neurdyna}. 
The absence of dynamics for $U$ can be interpreted as the special case when somatic capacitance is much smaller than the distal capacitance.
A spike is emitted when the membrane potential reaches a threshold $S_i(t) = \Theta(U_i(t))$, where $\Theta(x) = 0$ if $x < 0$, otherwise $1$ is the unit step function.
The constant $b_i$ represents the intrinsic excitability of the neuron.
The refractory mechanism is captured with the dynamics of $R_i$: the neuron inhibits itself after firing, by a constant weight $\rho$.
In contrast to standard \ac{IF} refractory mechanisms, a strong enough input can still induce the neuron to fire immediately after a spike.
The factors $\tau_{ref}$ and $\tau_{mem}$ are time constants of the membrane and refractory dynamics, respectively.
$I_{i}$ denotes the total synaptic current of neuron $i$, expressed as:
\begin{equation}\label{eq:Iv}
    \begin{split}
      \tau_{syn} \frac{\mathrm{d}}{\mathrm{d} t} I_{i}(t)  = & -I_{i}(t) + \sum_{j\in \text{pre}} W_{ij} S_j(t),
    \end{split}
\end{equation}
\noindent where $W_{ij}$ is the synaptic weights between pre-synaptic neuron $j$ and post-synaptic neuron $i$.
Because $V_i$ and $I_i$ are linear with respect to the weights $W_{ij}$, The dynamics of $V_i$ can be rewritten as:
\begin{equation}\label{eq:pq}
    \begin{split}
      V_i(t) =& \sum_{j\in \text{pre}} W_{ij} P_{ij}(t), \\
      \tau_{mem} \frac{\mathrm{d}}{\mathrm{d} t} P_{ij}(t)  = & -P_{ij}(t) +  Q_{ij}(t),\\
      \tau_{syn} \frac{\mathrm{d}}{\mathrm{d} t} Q_{ij}(t)  = & -Q_{ij}(t) +  S_{j}(t).
    \end{split}
\end{equation}
The states $P$ and $Q$ describe the traces of the membrane and the current-based synapse, respectively.
For each incoming spike, the trace $Q$ undergoes a jump of height $1$ and otherwise decays exponentially with a time constant $\tau_{\mathrm{syn}}$.
Weighting the trace $Q_{ij}$ with the synaptic weight $W_{ij}$ results in the \acp{PSP} of neuron $i$ caused by input neuron $j$.

All efferent synapses with identical time constants have identical dynamics.
By linearity of $P$ and $Q$, the state of the synapse can be described by a single synaptic variable per pre-synaptic neuron \cite{Brette_etal07_simunetw}.
In the equation above, this is evident by the fact that $P_{ij}$ and $Q_{ij}$ are only driven by $S_j$, and so the index $i$ can be dropped.
This results in as many $P$ and $Q$ variables as there are pre-synaptic neurons, independently of the number of synapses.
This strategy is commonly used in synapse circuits in neuromorphic hardware to reduce circuit area \cite{Bartolozzi_Indiveri06_silisyna}, and in software simulations of spiking neurons to improve memory consumption and computation time \cite{Brette_etal07_simunetw}.

\subsubsection*{Discrete Spike Response Model of the Neuron and Synapse Dynamics}

Because a computer will be used to simulate the dynamics, the dynamics are simulated in discrete time.
We denote the simulation time step with $\Delta t$.
We also make the layerwise organization of the network apparent with the superscript $l$ denoting the layer to which the neuron belongs.
The dynamical equations in \refeq{eq:clif} and \refeq{eq:pq} are expressed in discrete time as:
\begin{equation}\label{eq:lif_equations}
  \begin{split}
    U_i^l[t] &= \sum_j W^l_{ij} P_j^l[t] - \rho R^l_i[t]  + b^l_i, \\
    S_i^l[t] &= \Theta( U_i^l[t]), \\
  \end{split}
  \qquad
  \begin{split}
    P_j^l[t+\Delta t] &= \alpha P^l_{j}[t] + (1-\alpha) Q^l_{j}[t], \\
    Q_j^l[t+\Delta t] &= \beta  Q^l_{j}[t] + (1-\beta ) S^{l-1}_{j}[t], \\
    R_i^l[t+\Delta t] &= \gamma R^l_{i}[t] + (1-\gamma) S^{l}_{i}[t], \\
  \end{split}
\end{equation}
\noindent where the constants $\alpha=\exp(-\frac{\Delta t}{\tau_{\mathrm{mem}}})$, $\gamma=\exp(-\frac{\Delta t}{\tau_{\mathrm{ref}}})$ and $\beta=\exp(-\frac{\Delta t}{\tau_{\mathrm{syn}}})$ reflect the decay dynamics of the membrane potential $U$, the refractory state $R$ and the synaptic state $Q$ during a $\Delta t$ timestep.
Note that \refeq{eq:lif_equations} is equivalent to a discrete-time version of the \ac{SRM} with linear filters \cite{Gerstner_Kistler02_spikneur}.

\subsection{Deep Learning with Local Losses}\label{sec:decolle_derivation}
Loss functions are almost always defined using the network output at the top layer.
Assuming a global cost function $\mathcal{L}(S^N)$ defined on the spikes $S^N$ of the top layer and targets $\hat{Y}$, the gradients with respect to the weights in layer $l$ are:
\begin{equation}\label{eq:loss}
\frac{\partial \mathcal{L} (S^N)}{\partial W_{ij}^{l}} =
\frac{\partial \mathcal{L} (S^N)}{\partial S_{i}^{l}} 
\frac{\partial S^{l}_i          }{\partial U^{l}_{i}} 
\frac{\partial U^{l}_i          }{\partial W^{l}_{ij}}.
\end{equation}
The factor $\frac{\partial \mathcal{L} (S^N)}{\partial S_{i}^{l}}$ captures the backpropagated errors, \emph{i.e.} how changing the output of neuron $i$ in layer $l$ modifies the global loss.
This problem is known as the credit assignment problem.
It generally involves non-local terms, including the activity of other neurons, their errors, and their temporal history.
Thus, using local information only, a neuron in a deep layer cannot infer how a change in its activity will affect the top-layer cost.
An increasing body of work is showing that approximations to the backpropagated errors in \acp{SNN} can allow local learning, for example in feedback alignment \cite{Lillicrap_etal14_randfeed}. However, maintaining the history of the dynamics efficiently remains a challenging and open problem.
While it is possible to use \ac{BPTT} methods to compute these errors, this comes at a significant cost in memory and computation \cite{Williams_Zipser89_learalgo}, and is not consistent with the constraint of local information.


We address this conundrum using deep local learning \cite{Mostafa_etal17_deepsupe}.
We focus on a form of deep local learning that attaches random readouts to deep layers and defines auxiliary cost functions over the readout.
These auxiliary cost functions provide a task-relevant source of error for neurons in deep layers.
The random readout is obtained by multiplying the neural activations with a random and fixed matrix.
Training deep layers using auxiliary local errors that minimize the cost locally still allows the network as a whole to reach a small top-layer cost.
As explained in \cite{Mostafa_etal17_deepsupe}, minimizing a local readout's classification loss puts pressure on deep layers to learn useful task-relevant features, which allow the random local classifiers to solve the task.
Moreover, each layer builds on the features of the previous layer to learn features that are further disentangled with respect to the categories for its local random classifier compared to the previous layer.
Thus, even though no error information propagates downwards through the layer stack, the layers indirectly learn useful hierarchical features that end up minimizing the cost at the top layer.
Although the reasons for the effectiveness of local errors in deep network is intriguing and merits further work, it is orthogonal to the scope of this article. In this article, we focus on the fact that, provided local loss functions, surrogate learning in deep spiking neural networks becomes particularly efficient.

\subsection{\acf{DCLL}}\label{sec:dcll}
As discussed above, in \ac{DCLL}, we attach a random readout to each of the $N$ layers of spiking neurons:
\[ Y^l_i = \sum_j G^l_{ij} S_j^l, \]
where $G_{ij}^l$ are fixed, random matrices (one for each layer $l$) and $\Theta$ is an activation function.
The global loss function is then defined as the sum of the layerwise loss functions defined on the random readouts, \emph{i.e.} $\mathcal{L} = \sum_{l=1}^N L^l(Y^l)$.
To enforce locality, \ac{DCLL} sets to zero all non-local gradients, \emph{i.e.} $\frac{\partial L^l}{\partial W_{ij}^m} = 0$ if $m \ne l$.
With this assumption, the weight updates at each layer become:
\begin{equation}
  \Delta W_{ij}^l = - \eta \frac{\partial L^l}{\partial W_{ij}^l} = - \eta \frac{\partial L^l}{\partial S_{i}^{l}}  \frac{\partial S_i^l}{\partial W_{ij}^{l}} ,
\end{equation}
where $\eta$ is the learning rate.
Assuming the loss function depends only on variables in same time step, the first gradient term on the right hand side, $\frac{\partial L^l}{\partial S_{i}^{l}} $, can be trivially computed using the chain rule of derivatives.
Applying the chain of derivatives to the second gradient term yields:
\[
  \begin{split}
    \frac{\partial S_i^l}{\partial W_{ij}^{l}} & = \frac{\partial \Theta(U_i^l)}{\partial U_{i}^{l}} \frac{\partial U_i^l}{\partial W_{ij}^l}. \\
  \end{split}
\]
Due to the sparse, binary activation of spiking neurons, this expression vanishes everywhere except at 0, where it is infinite \cite{Neftci_etal19_surrgrad}.
To solve this problem, parameter updates in \ac{DCLL} are based on a differentiable but slightly different version of the task-performing network.
This approach was previously described as surrogate gradient-based learning \cite{Neftci_etal19_surrgrad,Zenke_Ganguli17_supesupe}:

\[
  \begin{split}
    \frac{\partial S_i^l}{\partial W_{ij}^{l}} &  = \sigma'(U_i^l) \frac{\partial U_i^l}{\partial W_{ij}^l},
  \end{split}
\]
where $\sigma'(U_i^l)$ is the surrogate gradient of the non-differentiable step function $\Theta(U_i^l)$.
The rightmost term is computed as:
\[
  \begin{split}
    \frac{\partial U_i^l}{\partial W_{ij}^l} & = P_j^l - \rho \frac{\partial R^l_i }{\partial W_{ij}^l}.
  \end{split}
\]
The terms involving $R^l_i$ are difficult to calculate because they depend on the spiking history of the neuron. As in Superspike, we ignore these dependencies and use regularization to favor low firing rates, a regime in which the $R_i^l$ has a negligible effect on the membrane dynamics.
Putting all three terms together, we obtain the \ac{DCLL} rule governing the synaptic weight update:
\begin{equation}
  \Delta W_{ij}^l = - \eta \frac{\partial L^l }{\partial S_{i}^{l}} \sigma'(U^l_i) P_j^l.
\end{equation}
In the special case of the Mean Square Error (MSE) loss for layer $l$, described as
\[
L^l = \frac{1}{2} \sum_{i} \left( Y^l_i-\hat{Y}^l_i \right)^2,
\]
the \ac{DCLL} rule becomes 
\begin{equation}\label{eq:dcll}
  \begin{split}
    \Delta W_{ij}^l & = - \eta\, \mathrm{error}_i^l\, \sigma'(U^l_i) P_j^l,\\
    \mathrm{error}_i^l &= \sum_j G_{ki}^{l} (Y^l_k-\hat{Y}^l_k), \\
  \end{split}
\end{equation}
\noindent
where $\hat{Y}^l$ is the pseudo-target for layer $l$.

\paragraph{Memory complexity of \ac{DCLL}:}
The variables $P$ and $U$ required for learning are local and readily available from the forward dynamics.
Because the errors are computed locally to each layer, \ac{DCLL} does not need to store any additional intermediate variables, \emph{i.e.} there is no space requirement for the parameter update computation.
The same neural traces $P$ and $Q$ maintained during the forward pass are sufficient (see \refsec{sec:autodiff}).
The computational cost of the weight update is the same as the Widrow-Hoff rule (one addition and two products per connection, see \refeq{eq:dcll}).
This makes \ac{DCLL} significantly cheaper to implement compared to \ac{BPTT} for training \ac{SNN}, \emph{e.g.} SLAYER \cite{Shrestha_Orchard18_slayspik} which scales spatially as $O(NT)$, where $T$ is the number of timesteps (see Appendix \refsec{sec:scaling} for details on scaling).

\paragraph{Sign-Concordant Feedback Alignment in the Local Layers}
The gradients of the local losses $L_i^l$ involve backpropagation through the local random projection $Y^l$.
This is a non-local operation as it requires the symmetric transpose of the matrix $G$.
This raises a weight transport problem, whereby the synaptic weight must be ``transported'' from one neuron to another.
In a von Neumann computer, this is not a problem since memory is shared across processes.
However, if memory is local, then a dedicated process must transmit this information.
Feedback alignment in non-spiking networks was demonstrated to overcome this problem at a cost in accuracy \cite{Mostafa_etal17_deepsupe}.
In our experiments, we use sign-concordant feedback weights to compute the gradients of the local losses: the backward weights have the same sign as the forward ones, but subject to fixed multiplicative Gaussian noise.
The noise here reflects the fact that weights do not need to be exactly symmetric.
This assumption is the most plausible scenario in mixed-signal neuromorphic devices, where connections can be programmed with the same sign bidirectionally, but the effective weights are subject to fabrication mismatch \cite{Neftci_etal11_systmeth}.
Since the weights in the local readouts are fixed, there is no weight transport problem during learning.
Thus, the computation of the errors can be carried out using another random matrix $H^l$ \cite{Lillicrap_etal16_randsyna} whose elements are equal to $H_{ij}^l = G_{ij}^{l,T}\omega_{ij}^l$ with a Gaussian distributed $\omega_{ij}^l \sim N(1,\frac12)$.
To enforce sign-concordance, all values $\omega_{ij}^l$ below zero were set to zero.

\paragraph{Biological Plausibility of \ac{DCLL} and Suitability for Neuromorphic Hardware:}
\refeq{eq:dcll} consists of three factors, one modulatory ($error_i$), one post-synaptic ($\sigma'(U_i^l)$) and one pre-synaptic ($P_j^l$).
These types of rules are often termed three-factor rules, which have been shown to be consistent with biology \cite{Pfister_etal06_optispik}, while being compatible with a wide number of unsupervised, supervised and reinforcement learning paradigms \cite{Urbanczik_Senn14_learby}.
The terms $P$ and $Q$ represent neural and synaptic states that are readily available at the neuron.
In our previous work and general experience, the shape of the surrogate function $\sigma$ does not play a major role in \ac{DCLL}\footnote{
  Conversely, the particular surrogate function is reported to play an important role in \ac{BPTT} \cite{Bellec_etal18_longshor}.
  This is likely due the product of the gradient approximations carried across multiple layers.
  This in turn can cause vanishing or exploding gradients.}.
The surrogate function $\sigma$ can be a piecewise linear function \cite{Neftci_etal17_evenranda}, such that $\sigma'$ becomes a boxcar function.
This corresponds to a learning update that is gated by the post-synaptic membrane potential, and is reminiscent of membrane voltage-based rules, where spike-driven plasticity is induced only when membrane voltage is inside an eligibility window \cite{Brader_etal07_learreal,Chicca_etal13_neurelec}.

In the derivation of the \ac{DCLL} rule, we used an instantaneous readout function $Y^l$ in the sense that it did not depend on states of the previous time step.
In biology, this readout would be carried out by spiking neurons.
This introduces a temporal dependency.
As in SuperSpike, this temporal dependency significantly increases the complexity of the learning, and is costly to implement in neuromorphic hardware.
One solution is to compute the errors using spiking neurons with dynamics faster than those of the hidden neurons.
In mixed signal hardware, this can be achieved through fast membrane and synaptic time constants. 
In digital hardware this could be achieved using a dedicated logic block.


\paragraph{Regularization and Implementation Details}
From a technological point of view, \acp{SNN} are interesting when the spike rate is low as dedicated neuromorphic hardware can directly exploit this sparsity to reduce computations by the same factor \cite{Merolla_etal14_millspik,Davies_etal18_loihneur}.
To ensure reasonable firing rates and prevent sustained firing, we use two regularizers.
One keeps $U$ below to the firing threshold on average, and one activity regularizer enforces a minimum firing rate in each layer. The final loss function is:
\begin{equation}
  \mathcal{L}_g = \sum_l L^l + \lambda_1 \langle [U_i^l + 0.01]^+ \rangle_{i} + \lambda_2 [0.1 - \langle U_i^l \rangle_{i}]^+
\end{equation}
where $\langle \cdot \rangle_i$ denotes averaging over index $i$, $[\cdot ]^+$ is a linear rectification, and $\lambda_1$, $\lambda_2$ are hyperparameters.
The minimum firing rate regularization is included to prevent the layers becoming completely silent during the training.
Our experiments used a piecewise linear surrogate activation function, such that its derivative becomes the boxcar function $\sigma'(x) = 1$ if  $x\in [-.5, .5]$ and $0$ otherwise. 

In all our experiments, weight updates are made for each time step of the simulation.
We use the AdaMax optimizer \cite{Kingma_Ba14_adammeth} with parameters $\beta_1=0$, $\beta_2=95$ and learning rate $10^{-9}$, and a smooth L1 loss.
Biases were used for all layers and trained in all \ac{DCLL} layers.
The weights $G^l$ used for the local readouts were initialized uniformly.
PyTorch code and a tutorial are publicly available on Github\footnote{\url{https://github.com/nmi-lab/decolle-public}}.
\ac{DCLL} is simulated using mini-batches to leverage the GPU's parallelism.

\subsection{Computational Graph and Implementation using Automatic Differentiation:}\label{sec:autodiff}

Perhaps one of the strongest advantages of \ac{DCLL} is its out-of-the-box compatibility with \ac{AD} tools for implementing gradient \ac{BP}.
\ac{AD} is a technology recently incorporated in machine learning frameworks to automatically compute gradients in a computational graph\footnote{Gradient \ac{BP} is a special case of reverse mode \ac{AD}, see \cite{Baydin_etal17_autodiff} for complete review}.
\ac{AD} operates on the principle that all numerical computations are compositions of a finite set of elementary operations for which derivatives are known. 
By combining the derivatives of the operations through the chain rule of derivatives, the derivative of the overall composition can be computed in a single pass \cite{Baydin_etal17_autodiff}.

In practice, machine learning frameworks augment each elementary computation with its corresponding derivative function. As the desired operation is constructed, the dependencies with other variables are recorded as a computational graph. To perform gradient \ac{BP}, after a forward pass, a backward pass computes all the derivatives of the operations in the graph. The root node of the reverse graph is typically a scalar loss function, and the leaf nodes are generally inputs and parameters of the network. After the backward pass, the gradients of all leaf nodes are applied to the trained parameters or inputs according to the optimization routine (\emph{e.g.} Adam or similar).

\begin{SCfigure}
    \includegraphics[width=.42\textwidth]{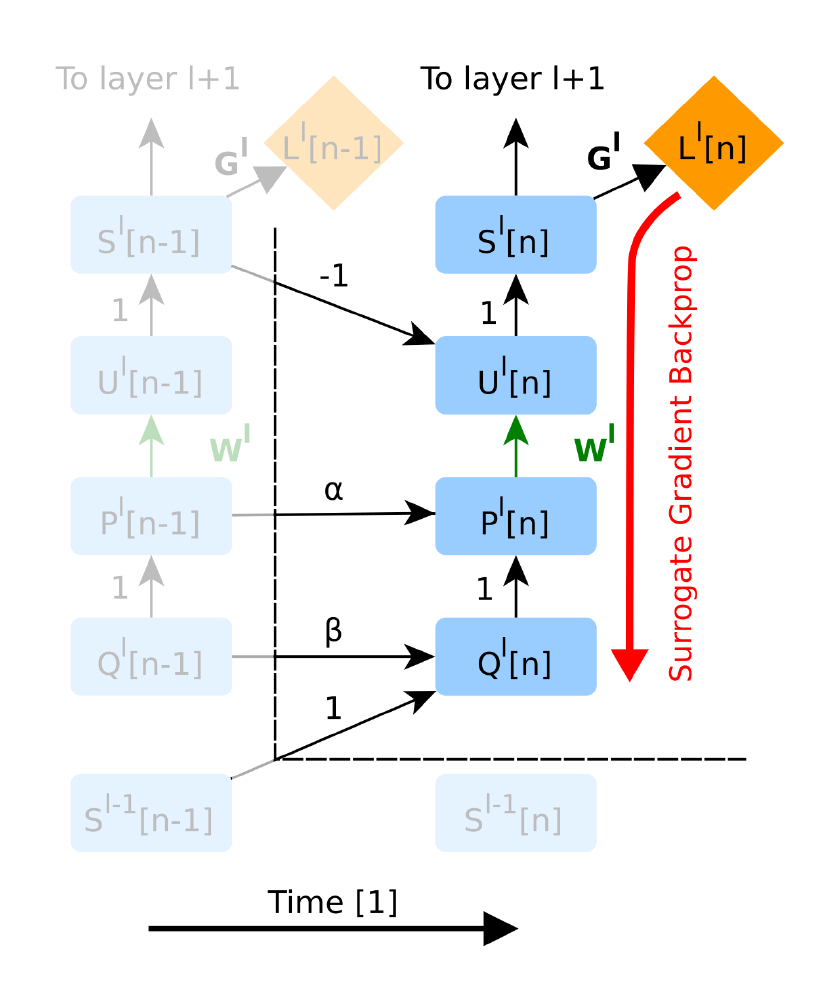}
  \caption{\label{fig:cg_dcll} The unfolded computational graph of a feedforward \ac{SNN}. Time flows to the right. Only temporal dependencies between timestep $n-1$ and $n$ are shown here. Green edges indicate variables trained in the presented version of \ac{DCLL}. Red edges indicate the flow of the gradients. Note that this graph is similar to that of a simple recurrent neural network. The forward \ac{RTRL} approach combined with local errors means that errors do not propagate through neurons and across layers, as all the information required for learning is available at the layer and the current time step $n$. For implementation purposes however, autodifferentiation can be used to compute gradients within the neuron and time step (see Appendix \refsec{sec:autodiff} for details). To avoid clutter, the the node for $R$ has been omitted. }
\end{SCfigure}

\acp{SNN} being a special case of recurrent neural networks, it is possible to apply \ac{AD} to the full graph \cite{Shrestha_Orchard18_slayspik}. 
On the other hand, \ac{DCLL} only requires backpropagating through a subgraph corresponding to one layer and within the same time step \reffig{fig:cg_dcll}.
This is because the information necessary for computing the gradients ($P$, $Q$, $R$, and $U$) is carried forward in time, and because local loss functions provide gradients for each layer.

\ac{AD} in \ac{DCLL} thus computes the gradients, locally, for each layer within each timestep.
Because some operations in the subgraph can be non-differentiable (such as the spiking nonlinearity), we call this the ``surrogate gradient backprop'' \reffig{fig:cg_dcll}.
This integration allows leveraging the layers, operations, optimizers and cost functions provided by the software.
All experiments under the Experiments section use \ac{AD} to compute derivatives. 
To prevent \ac{AD} from unnecessarily backpropagating in time, we rely on special ``stop-gradient'' operations.
In the appendix, we provide pseudocode and discussion of how this can be achieved.

\section{Experiments}
\subsection{Regression with Poisson Spike Trains}
To illustrate the inner workings of \ac{DCLL}, we first demonstrate \ac{DCLL} in a regression task.
A three-layer fully connected network consisting of 512 neurons each is stimulated with a fixed $500$ms Poisson spike train.
Each layer in the network is trained with a different pseudo-target: $\hat{Y}^1$, a ramp function; $\hat{Y}^2$, a high-frequency sinusoidal function and $\hat{Y}^3$, a low-frequency sinusoidal function.  \reffig{fig:dcll_illustration} illustrates the states of the neurons.
For illustration purposes, the recording of the neural states was made in the absence of parameter updates (\emph{i.e.} the learning rate is 0).
The refractory mechanism decreases the membrane potential after the neuron spikes ($U[t]$).
As discussed in the methods we use regularization on the membrane potential to keep the neurons from sustaining high firing rates and an activity regularizer to maintain a minimum firing rate.
Updates to the weight are made at each time step and can be non-zero when the derivative of the activation function $\sigma'(U)$ and $P$ are non-zero.
The magnitude and direction of the update are determined by the error.
Note that, in effect, the error is randomized as a consequence of the random local readout.
The network learned to use the input spike times to reliably produce the targets.

\subsection{N-MNIST}
The N-MNIST dataset was recorded with a \ac{DVS} \cite{Lichtsteiner_etal08_128x120d} mounted on a pan-tilt unit performing microsaccadic motions in front of a screen displaying samples from the MNIST dataset \cite{Orchard_etal15_convstat}.
Unlike standard imagers, the \ac{DVS} records streams of events that signal the temporal intensity changes at each of its $128 \times 128$ pixels.
For each of the 10 digits, we used 2000 samples for training and 100 samples for testing.
The samples are cropped spatially from $34\times 34$ to $32\times 32$ and temporally to $300$ms for both the train and test set.
The network is simulated with a $1$ms resolution -- in other words, we sum up events in $1$ms time bins.
No further pre-processing is applied to the events.

Events are separated in two channels with respect to their polarity.
A N-MNIST sample is therefore represented as a tensor of shape $300 \times 2 \times 32 \times 32$, stacked into mini-batch of 500 samples.
The \ac{DCLL} network is fed with $1$ms slices of the input at a time.
We relied on the same three-layer convolutional architecture used in the DvsGesture task described below.
After a ``burn-in'' period of $50$ms during which no update is made, gradient updates are performed at every simulation step.
Hence, there are 250 weight updates per mini-batch.
While the relevance of the time domain in N-MNIST is debatable \cite{Iyer_etal18_neurmnis}, this experiment shows that the neural dynamics of our network leads to successful classifications in under $300$ms.

\begin{figure}
\centering
\includegraphics[height=.20\textheight]{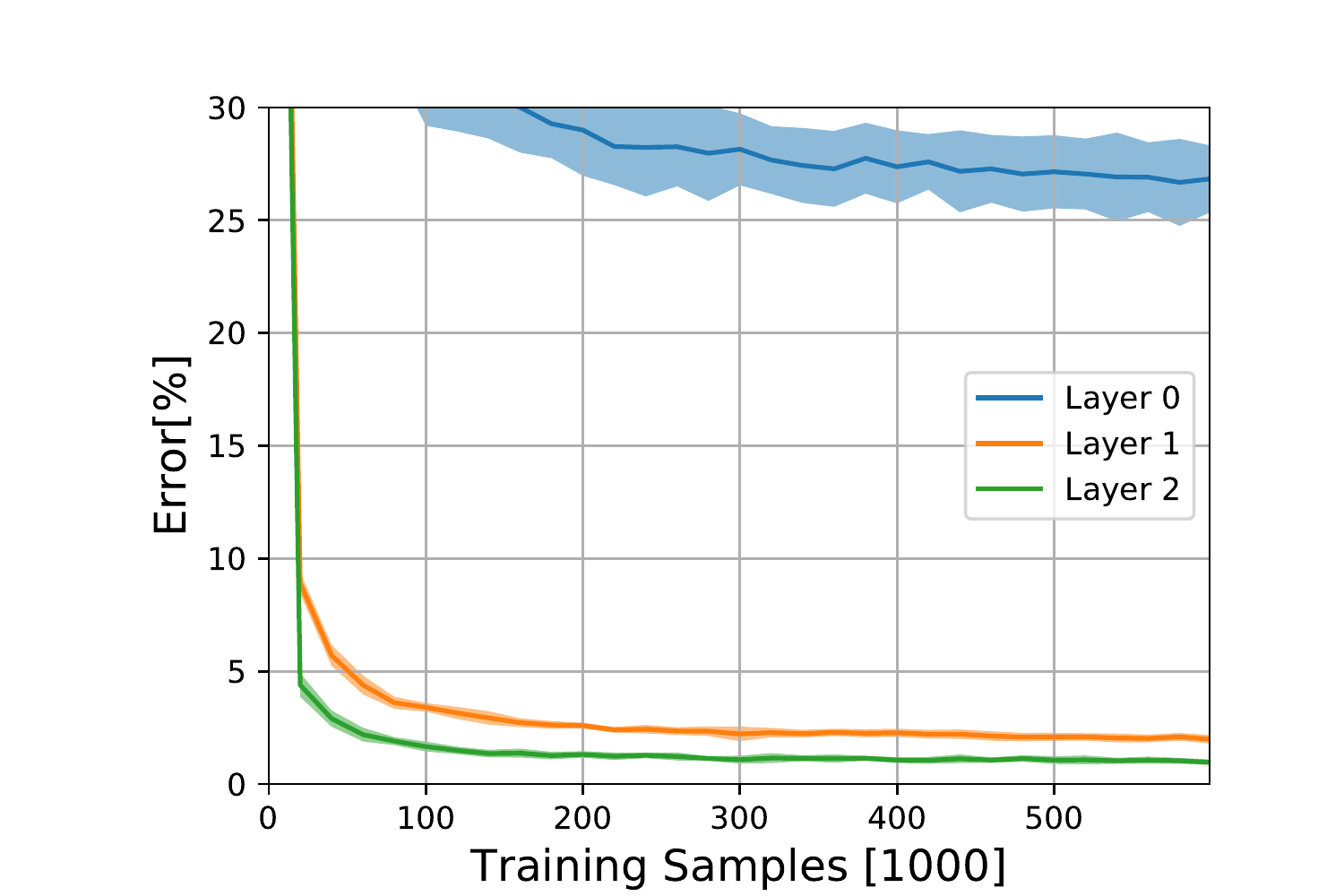}
  \caption{
    Classification results on the N-MNIST dataset for the three DECOLLE layers.
    Classification Error for the N-MNIST task during learning for all local errors associated with the convolutional layers.
    Shadings indicate standard deviation across the 10 runs.
  }
  \label{fig:n-mnist}
\end{figure}

The results on the N-MNIST dataset are shown in Figure \ref{fig:n-mnist}.
The experiment was performed 10 times with different random seeds.
\ac{DCLL}'s final error is $0.96 \% \pm 0.12\%$  for the third layer with 600 000 training iterations.
We note that, due to the large memory requirements, it is not practical to train the \ac{DCLL} convolutional network using \ac{BPTT}.
Hence we cannot provide \ac{BPTT} baselines.

\subsection{DvsGesture}
\begin{figure}
\centering\includegraphics[height=.15\textheight]{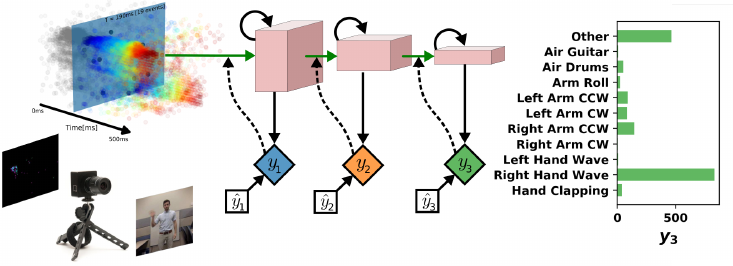}
\centering\includegraphics[height=.15\textheight]{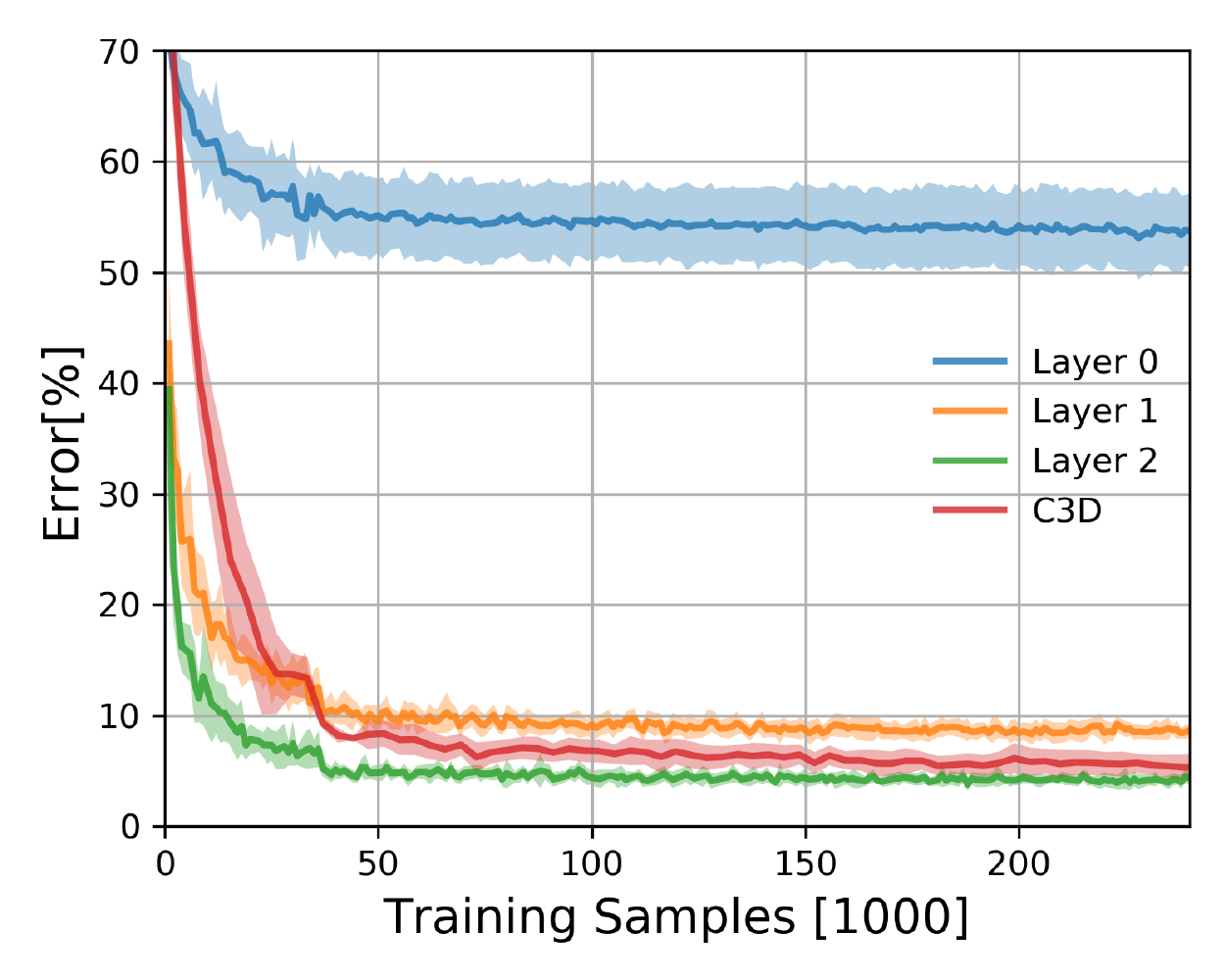}
\caption{(Left) \ac{DCLL} setup for DvsGesture recognition. Learning was performed on the dataset provided with \cite{Amir_etal17_lowpowe} and consists of 11 gestures. The network consisted of three convolutional layers with max-pooling. A local classifier is attached to every layer and followed by dropout for regularization. \ac{DCLL} is fed with $1$ms integer frames. (Right) Classification Error for the DvsGesture task during learning for all local errors associated with the convolutional layers \label{fig:convergence_dvs_gestures}. Shadings indicate standard deviation across runs (5 runs for C3D, 10 runs for \ac{DCLL})}
\end{figure}

We test \ac{DCLL} at the more challenging task of learning gestures recorded using a \ac{DVS}.
Amir \emph{et al.} recorded the DvsGesture dataset using a DVS, which comprises 1342 instances of a set of 11 hand and arm gestures, collected from 29 subjects under 3 different lighting conditions \cite{Amir_etal17_lowpowe}.
The unique features of each gesture are embedded in the stream of events.
The event streams were downsized from $128\times 128$ to $32\times 32$ (events from 4 neighboring pixels were summed together as a common stream) and binned in frames of $1\mathrm{ms}$, the effective time step of the GPU-based simulation \reffig{fig:convergence_dvs_gestures}.
During training, a sample consisted of $500$ms-long slices of the sample.
To maximize the use of the dataset, the starting point of the slice was picked randomly, but such that a full $500$ms sequence could be constructed.
The sequences were presented to the network in mini-batches of 72 samples.
Testing sequences were $1800$ms-long, each starting from the beginning of each recording (288 testing sequences).
Note that since the shortest recording in the test set is $1800$ms, this duration was selected to simplify and speed up the evaluation.
The classification is obtained by counting spikes at the output starting from a burn-in period of $50$ms and selecting as output class the neuron that spiked the most. Test results from the \ac{DCLL} network are reported with the dropout layer kept active, as this provided better results.
Contrary to \cite{Amir_etal17_lowpowe}, we did not use stochastic decay and the neural network structure is a three-layer convolutional neural network, loosely adapted from \cite{Springenberg_etal14_strisimp}.
We did not observe significant improvement by adding more than three convolutional layers.
In shallow convolutional neural networks, it is common to use larger kernel sizes \cite{Kubilius_etal19_braiobje,LeCun_etal98_gradlear}. Since the input sizes were 32x32, we used 7x7 kernel sizes in \ac{DCLL} to ensure that the receptive field of neurons in the last layer covered the input.
The optimal hyperparameters were found by a combination of manual and grid search.
The learning rate was divided by 5 every 500 steps.

We compared with C3D and energy-efficient deep networks (EEDN).
EEDN is a convolutional deep neural network architecture that can be trained offline (\emph{e.g.} on a GPU) and deployed on the IBM TrueNorth chip \cite{Esser_etal16_convnetw}. 
EEDN was applied to DVS gestures and provides an important benchmark on this task \cite{Amir_etal17_lowpowe}.
Because EEDN was not designed to utilize the temporal dynamics of the spiking neurons in IBM TrueNorth chip, time is represented using the channel dimension of 2D convolutional networks.
This approach limits the length of the sequence that EEDN can process. 
To overcome this, \cite{Amir_etal17_lowpowe} used a sliding window filter. 
C3D is a 3D convolutional network commonly used for spatiotemporal classification in videos \cite{Tran_etal15_learspat}, where the dimensions are time, height and width.
Using 3D kernels, C3C can learn spatiotemporal patterns.
The network was similar to \cite{Tran_etal15_learspat} except that is was adapted for 32x32 frames and using half of the features per layer (see Appendix for network layers). We note that the C3D network is deeper and wider than the \ac{DCLL} network.
We found that 16x32x32 frames, where each of the 16 representing 32ms slices of the DVS data performed best.

\begin{SCfigure}
\centering\includegraphics[height=.15\textheight]{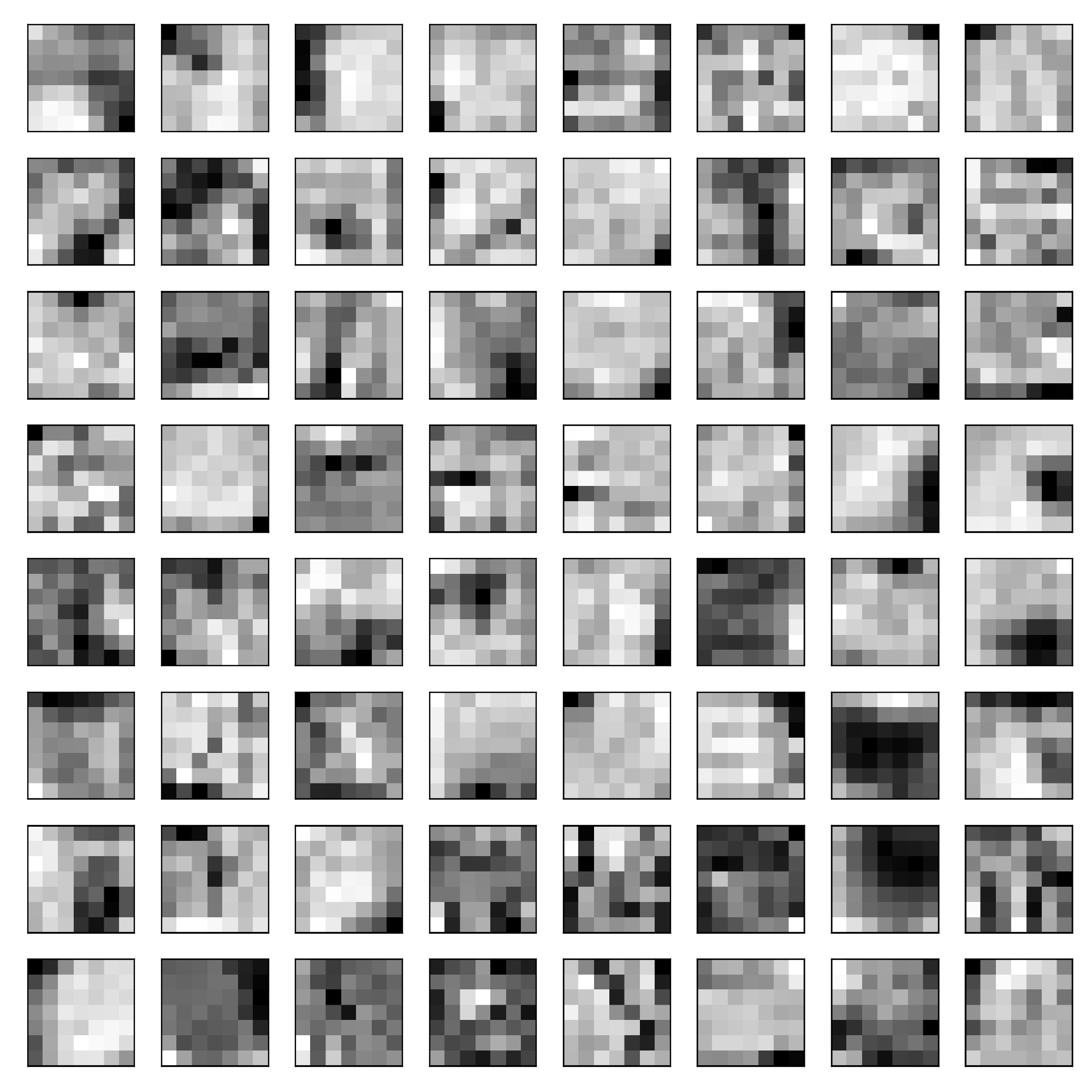}
\hspace{.2pt}
\centering\includegraphics[height=.15\textheight]{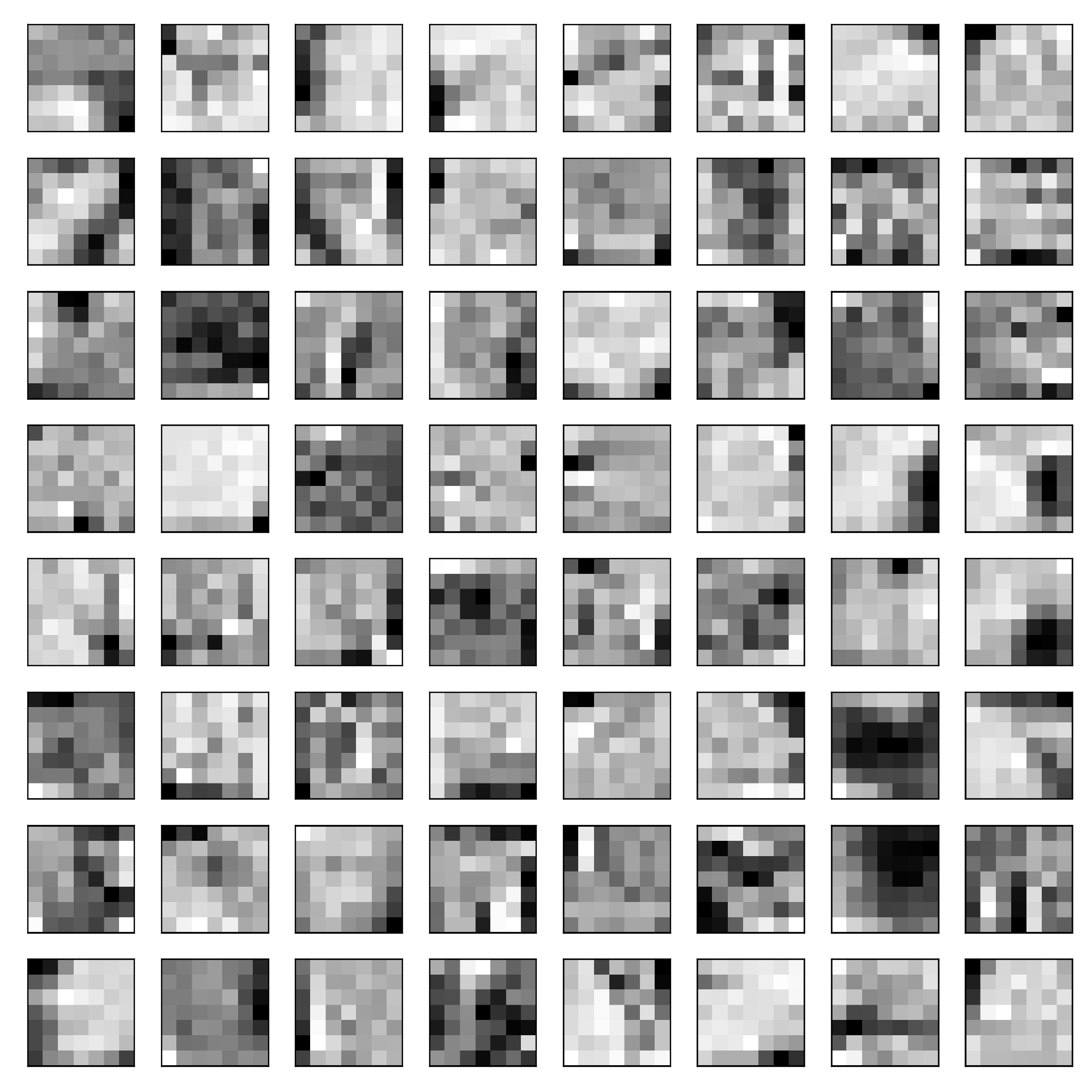}
\caption{ $7\times 7$ Filters learned in the positive polarity channel (Left) and negative polarity channel (Right) of the first convolutional layer. The similarity of the kernels across the two polarities reflects the DVS data, where leading edges and trailing edges co-occur with opposite polarities. \label{fig:kernels}}
\end{SCfigure}

Overall, \ac{DCLL}'s performance is comparable or better than other published \ac{SNN} implementations that use \ac{BP} for training (\reftab{tab:dvsgestures_accuracy}, \reffig{fig:convergence_dvs_gestures}) and close to much larger C3D networks.
\ac{DCLL} reached the reported accuracies after two orders of magnitude fewer iterations and smaller network compared to the IBM EEDN case \reftab{tab:dvsgestures_accuracy}, \cite{Amir_etal17_lowpowe}.
\begin{table}
\begin{center}
\begin{tabular}{l l l l l}
  \textbf{Model} & \textbf{Error} & \textbf{Training} & \textbf{Iterations} & \textbf{Ref.}\\
     \ac{DCLL}& $\mathbf{4.46\pm 0.16\%}$ & \textbf{Online} & $\mathbf{.16}$\textbf{M} &This Work   \\
     SLAYER & $6.36\pm 0.49$ \% & BPTT & $.27$M & \cite{Shrestha_Orchard18_slayspik}  \\
     C3D  & $5.46\pm 1.06\%$ & BPTT  &$.32$M & \cite{Tran_etal15_learspat}  \\
     IBM EEDN  & $8.23\%$ ($5.51\%$) & BPTT & $64$M & \cite{Amir_etal17_lowpowe} \\
\end{tabular}\\
\end{center}
\caption{\label{tab:dvsgestures_accuracy} Classification error at the DvsGesture task. The IBM EEDN error in parentheses refers to the case with sliding window filter.
}
\end{table}
\begin{SCtable}
  \scriptsize
    \begin{tabular}{lllc}
      Layer Type              & \#  & Data Type & Dimensions \\
      \hline
      DVS                         & 2  &AEDAT 3.1 &$128\times 128$\\
      Downsample (Sum)            & 2  &Integer    & $32\times 32$ \\
      $7 \times 7$ Conv           & 64 &Float    & $30\times 30$ \\
      $2 \times 2$ MaxPool        & 64 &Float    & $15\times 15$ \\
      Spiking Non-linearity & &Binary & \\
      $\quad \quad$ Dropout(p=.5) &    &Float     &               \\
      $\quad \quad$ Dense         & 11 &Float     & $11$          \\
      $7 \times 7$ Conv           & 128&Float    & $13\times 13$ \\
      Spiking Non-linearity & &Binary & \\
      $\quad \quad$ Dropout(p=.5) &    &Binary     &               \\
      $\quad \quad$ Dense         & 11 &Float     & $11$          \\
      $7 \times 7$ Conv           & 128&Float    & $11\times 11$ \\
      $2 \times 2$ MaxPool        & 128&Float    & $5\times 5$   \\
      Spiking Non-linearity & &Binary & \\
      $\quad \quad$ Dropout(p=.5) &    &Binary     &               \\
      $\quad \quad$ Dense         & 11 &Float     & $11$          \\
      \hline
    \end{tabular}
  \caption{\label{tab:allcnn} \ac{DCLL} Neural network used for the DvsGesture dataset. Note that dense layers are used for the local classifiers only and were not fed to the subsequent convolutional layers. AEDAT 3.1 is a data format used for event-based data. The spiking nonlinearity was always applied after the pooling layers. Dropout layers were left active during testing.}
\end{SCtable}
Interestingly, the first layer of \ac{DCLL} has a low classification accuracy.
A similar effect is observed in non-spiking neural networks \cite{Mostafa_etal17_deepsupe}.
The local classifier errors improve for the second and third hidden layers compared to the first hidden layer.
This is an indication that the network is able to make use of depth to obtain better accuracy.
An examination of the filters learning in the first convolutional layer shows filters of varying frequencies and orientations \reffig{fig:kernels}. 
Interstingly, the filters on the positive and negative channels of the DVS are similar, but exhibit small variations that are consistent with motion. 
This correlation is consistent with the DVS data, where leading edges of one polarity co-occur with trailing edges of opposite polarity.

\section{Conclusion}
Understanding and deriving neural and synaptic plasticity rules that can enable hidden weights to learn is an ongoing quest in neuroscience and neuromorphic engineering.
From a machine learning perspective, locality and differentiability are key issues of the spiking neuron model operations. While the latter problem is now being tackled with surrogate gradient approaches, how to achieve this in deep networks in a scalable and local fashion is still an open question.

We presented a novel synaptic plasticity rule, \ac{DCLL}, derived from a surrogate gradient approach with linear scaling in the number of neurons. The rule draws on recent work in surrogate gradient descent in spiking neurons and local learning with layerwise classifiers.
The linear scalability is obtained through a (instantaneous) rate-based cost function on the local classifier.
The simplicity of the \ac{DCLL} rule equation makes it amenable for direct exploitation of existing machine learning software libraries. 
Thanks to the surrogate gradient approach, the updates computed through automatic differentiation are equal to the \ac{DCLL} update.
This enables the leveraging of a wide variety of machine learning frameworks for implementing online learning of \acp{SNN}.

Updates in \ac{DCLL} are performed at every time step, in accordance with the continuity of the leaky \ac{IF} dynamics. This can lead to a large number of updates and inefficient implementations in hardware. To tackle this problem, updates can be made in an error-triggered fashion, as discussed in \cite{Payvand_etal20_errothre}.
A direct consequence of the local classifiers is the lack of cross-layer adaptation of the layers.
To tackle this problem, one could use meta-learning to adapt the random matrix in the classifier.
In effect, the meta-learning loop would act as the outer loop in the synthetic gradients approach \cite{Jaderberg_etal16_deconeur}.
The notion that a ``layer'' of neurons specialized in solving certain problems and sensory modalities is natural in computational neurosciences and can open multiple investigation avenues for understanding learning and plasticity in the brain.

\ac{DCLL} is a departure from standard \acp{SNN} trained with Hebbian spike-timing-dependent plasticity, as it uses a normative learning rule that is partially derived from first principles.
Models of this type can make use of standard processors where it makes the most sense (i.e. readout, cost functions etc.) and neuromorphic dedicated hardware for the rest.
Because it leverages the best of both worlds, \ac{DCLL} is poised to make \ac{SNN}s take off in event-based computer vision.

\subsubsection*{Acknowledgments}
EN was supported by the Intel Corporation, the National Science Foundation under grant 1652159, and by the Korean Institute of Science and Technology.
JK was supported by a scholarship within the FITweltweit programme of the German Academic Exchange Service (DAAD).
HM was supported by the Swiss National Fund.
JK, HM, EN wrote the paper, conceived the experiments.
JK and EN ran the experiments and analyzed the data.
The authors declare that this study received funding from Intel Corporation.
The funder was not involved in the study design, collection, analysis, interpretation of data, the writing of this article or the decision to submit it for publication.


\bibliographystyle{plain}
\bibliography{biblio_unique_alt}
\pagebreak
\section{Supplementary Information}

\subsection{Implementation of DECOLLE using Autodifferentiation}\label{sec:autodiff_appendix}
The equations of \ac{DCLL} are very similar to that of a simple recurrent neural network. However, rather than performing backpropagation through-time, the derivatives of $U_i$ are computed by propagating the traces $P_i$ forward in time as follows:

\begin{algorithm}[H]
\begin{algorithmic}
\FOR {$n = 0 ... T$}
\STATE \COMMENT{Advance State}
\FOR{$ l<1 ... L   $}
\STATE $U_{i}^l = \sum_j W_{ij}^l P_{j}^l - \rho R_i^l$
\STATE $S_i = STOPGRAD(\Theta(U_i) -\sigma(U_i)) + \sigma(U_i)$
\IF{Sign-concordant feedback matrix}
\STATE $Y_i = STOPGRAD(\sum_j G_{ij} S_j - \sum_j H_{ij} S_j) + \sum_j H_{ij} S_j$
\ELSE
\STATE $Y_i = \sum_j G_{ij} S_j$
\ENDIF
\STATE $L^l = f_{loss}(Y_k^l, \hat{Y}_k^l)$
\STATE $W_{ij}^l = W_{ij}^l + \eta GRAD(L^l, W_{ij}^l)$
\STATE $P_{i}^l = \alpha P_{i}^l + (1-\alpha_i) Q_{i}^l$
\STATE $Q_{i}^l = \beta Q_{i}^l  + (1-\beta_i)  STOPGRAD(S_{i}^{l-1})$
\STATE $R_{i}^l = \gamma R_{i}^l + (1-\gamma_i) STOPGRAD(S_{i}^{l-1})$
\ENDFOR
\ENDFOR
\end{algorithmic}
\label{alg:seq}
\end{algorithm}

where $f_{loss}$ is the loss function, \emph{e.g.} MSE loss.
STOPGRAD prevents the flow of gradients by setting them to zero. STOPGRAD(A-B)+B as above is a common construct used to compute gradients using a separate subgraph. In this case, it is used to implement the surrogate gradient.
Note that the time variable does not appear in the variables.
This underlines that \ac{DCLL} computation does not need to store the state histories, and that the variables necessary for computing the gradient (GRAD) are available within the same step.
In our implementation, the weights are updated online, at every time step.
The cost of making the parameter update is no more than accumulating the gradients at each time step, since parameter memory and dynamical state memory are the same.
Therefore, there is no overhead in updating online.
Note that this may not be the case in dedicated neuromorphic hardware or AI accelerators that require different memory structures for storing parameter memory.

\subsection{Complexity Overhead for Various Spiking Neuron Gradient-Based Training Approaches} \label{sec:scaling}
We provide additional detail on the complexity of \ac{DCLL} compared to other learning methods.
In the current implementation of \ac{DCLL}, all weight updates are applied immediately, thus no additional memory is necessary to accumulate the gradients. In all other methods presented in \reftab{tab:complexity_analysis}, the weight updates are applied in an epoch-wise fashion, which requires an additional variable to store the accumulated weights. However, this is an implementation choice which could have been made for methods other than DECOLLE as well. For this reason, the overhead of accumulating gradients in epoch-wise learning is ignored in the following calculations.
\paragraph{DECOLLE}
The state of $P$ and $Q$ must be maintained.
These states are readily available from the forward pass, and therefore do not need to be stored specifically for learning.
Space complexity is therefore $O(1)$.
Each weight update requires $M N_r$ multiplications to obtain $M$ local errors.
Each of these are multiplied by the number of inputs $pN$, resulting in $O(pNM + M N_r)$ time complexity, where $p$ is the faction of connected neurons.
Similarly to \cite{Mostafa_etal17_deepsupe}, the random weights in $G^l$ can be computed using a random number generator, which requires one seed value per layer.
\paragraph{Superspike} When using the Van Rossum Distance (VRD), the Superspike learning rule requires one trace per connection, resulting in a space complexity of $O(pNM)$.
The additional complexity here compared to DECOLLE is caused by the additional filter in the Van Rossum distance.
Note that if the learning is applied directly to membrane potentials, the space and time complexity is similar to that of DECOLLE.
\paragraph{eProp} In the case when no future errors are used, the complexity of e-Prop \cite{Bellec_etal19_biolinsp} is similar to that of SuperSpike.
\paragraph{RTRL and BPTT} The complexity of these techniques are discussed in detail in \cite{Williams_Zipser95_gradlear}.

\begin{table}[h!]
  \centering
    \begin{tabular}{lll}
      Method          & Space            & Time             \\ \hline
      DECOLLE         & $O(1)$           & $O(M N_r + pNM)$ \\
      SuperSpike (VRD)& $O(pNM)$         & $O(pNM)$       \\
      e-Prop 1        & $O(pNM)$         & $O(pNM)$      \\
      RTRL            & $O(pNM^2)$       & $O(p^2 N^2 M^2)$  \\
      BPTT            & $O(N T)$         & $O(pNMT)$        \\
      \hline
    \end{tabular}
    \caption{\label{tab:complexity_analysis} Complexity analysis of the gradient computation. $N$: Input neurons, $M$: Neurons in Layer, $T$: Length of Backpropagated Sequence, $N_r$: number of readout neurons in DECOLLE, $p$: ratio of connected neurons/total possible connections.}
\end{table}

\subsection{C3D Network}
We used a standard 3D convolution network (C3D) for comparison with \ac{DCLL}. In C3D, the temporal dimension is taken into account in the third dimension of the 3D convolution.
\begin{center}
    \begin{tabular}{lllc}
      Layer Type              & \#  & Data Type & Dimensions \\
      \hline
      DVS                              & 2    &AEDAT 3.1 & $128\times 128$\\
      Downsample (Sum) and Frame       & 2    &Binary    & $16 \times 32\times 32$ \\
      $3 \times 3 \time  3$ Conv ReLU  & 32   &Binary    & $16 \times 32\times 32$ \\
      $1 \times 2 \times 2$ MaxPool    & 32   &Binary    & $16\times 16 \times 16$ \\
      $3 \times 3 \time  3$ Conv ReLU  & 64   &Binary    & $16\times 16 \times 16$ \\
      $2 \times 2 \times 2$ MaxPool    & 64   &Binary    & $8 \times 8 \times 8$ \\
      $3 \times 3 \time  3$ Conv ReLU  & 128  &Binary    & $8 \times 8 \times 8$ \\
      $3 \times 3 \time  3$ Conv ReLU  & 128  &Binary    & $8 \times 8 \times 8$ \\
      $2 \times 2 \times 2$ MaxPool    & 128  &Binary    & $4 \times 4 \times 4$ \\
      $3 \times 3 \time  3$ Conv ReLU  & 256  &Binary    & $4 \times 4 \times 4$ \\
      $3 \times 3 \time  3$ Conv ReLU  & 256  &Binary    & $4 \times 4 \times 4$ \\
      $2 \times 2 \times 2$ MaxPool    & 256  &Binary    & $2 \times 2 \times 2$ \\
      $3 \times 3 \time  3$ Conv ReLU  & 256  &Binary    & $2 \times 2 \times 2$ \\
      $3 \times 3 \time  3$ Conv ReLU  & 256  &Binary    & $2 \times 2 \times 2$ \\
      $2 \times 2 \times 2$ MaxPool    & 256  &Binary    & $1 \times 2 \times 2$ \\
      Dense  ReLU                      & 1024 &Float     & $1024$          \\
      Dense  ReLU                      & 512  &Float     & $512$          \\
      Dense Softmax        & 11 &Float     & $11$          \\
      \hline
    \end{tabular}
\end{center}

\end{document}